\documentclass[letterpaper]{article} 
\usepackage[submission]{aaai2026}  
\usepackage{times}  
\usepackage{helvet}  
\usepackage{courier}  
\usepackage[hyphens]{url}  
\usepackage{graphicx} 
\usepackage{natbib}  
\usepackage{caption} 
\pdfoutput=1

\usepackage{tabularray}
\UseTblrLibrary{booktabs}
\usepackage{amsmath}

\usepackage{algorithm}
\usepackage{algorithmic}

\usepackage{newfloat}
\usepackage{listings}
\DeclareCaptionStyle{ruled}{labelfont=normalfont,labelsep=colon,strut=off} 
\floatstyle{ruled}
\newfloat{listing}{tb}{lst}{}
\floatname{listing}{Listing}
\makeatletter
\def\@fnsymbol#1{\ensuremath{\ifcase#1\or \dagger\or \ddagger\or \mathsection\or \mathparagraph\or \|\or **\or \dagger\dagger \or \ddagger\ddagger \else\@ctrerr\fi}}
\makeatother

\title{Text-guided Controllable Diffusion for Realistic Camouflage Images Generation}
\author{
    Yuhang Qian\textsuperscript{\rm 1,2},   
    Haiyan Chen\textsuperscript{\rm 2},
    Wentong Li\textsuperscript{\rm 1},
    Ningzhong Liu\textsuperscript{\rm 2},
    Jie Qin\textsuperscript{\rm 1}\thanks{Corresponding Author.}
}
\affiliations{
    \textsuperscript{\rm 1} MoE Key Laboratory of Brain-Machine Intelligence Technology, College of Artificial Intelligence, Nanjing University of Aeronautics and Astronautics, Nanjing 211106, Jiangsu, China\\
    \textsuperscript{\rm 2} College of Computer Science and Technology, Nanjing University of Aeronautics and Astronautics, Nanjing 211106, Jiangsu, China

    qianyuhang@nuaa.edu.cn
}

\usepackage{bibentry}

\begin{document}

\maketitle

\begin{abstract}
Camouflage Images Generation (CIG) is an emerging research area that focuses on synthesizing images in which objects are harmoniously blended and exhibit high visual consistency with their surroundings. 
Existing methods perform CIG by either fusing objects into specific backgrounds or outpainting the surroundings via foreground object-guided diffusion. 
However, they often fail to obtain natural results because they overlook the logical relationship between camouflaged objects and background environments.
To address this issue, we propose CT-CIG, a Controllable Text-guided Camouflage Images Generation method that produces realistic and logically plausible camouflage images.
Leveraging Large Visual Language Models (VLM), we design a Camouflage-Revealing Dialogue Mechanism (CRDM) to annotate existing camouflage datasets with high-quality text prompts.
Subsequently, the constructed image-prompt pairs are utilized to finetune Stable Diffusion, incorporating a lightweight controller to guide the location and shape of camouflaged objects for enhanced camouflage scene fitness. 
Moreover, we design a Frequency Interaction Refinement Module (FIRM) to capture high-frequency texture features, facilitating the learning of complex camouflage patterns. 
Extensive experiments, including CLIPScore evaluation and camouflage effectiveness assessment, demonstrate the semantic alignment of our generated text prompts and CT-CIG's ability to produce photorealistic camouflage images. 
\end{abstract}

\begin{links}
    \link{Code}{https://github.com/NikoNairre/CT-CIG}
\end{links}

\section{Introduction}

Camouflage is an instinctive survival mechanism that organisms utilize to blend into their surroundings, making them visually indistinct, either to evade predators or to ambush prey \cite{merilaita2017camouflage}. 
Owing to its interesting and challenging characteristics, camouflage vision perception has gained attention in research such as Camouflaged Object Detection (COD) \cite{fan2020camouflaged, luo2024vscode, yan2025ucod} and Concealed Instance Ranking (CIR) \cite{lv2021simultaneously}.
However, progress in this field is hampered by a performance bottleneck stemming from the scarcity of camouflage training datasets, underscoring the requirement for effective data acquisition methods.
The difficulty of collecting natural camouflage images—owing to intricate and environment-specific conditions—has spurred the emergence of Camouflage Images Generation (CIG) \cite{zhao2025deep}, a paradigm focused on synthesizing artificial camouflage images.
\begin{figure}[t]
    \centering
    \includegraphics[width=\linewidth]{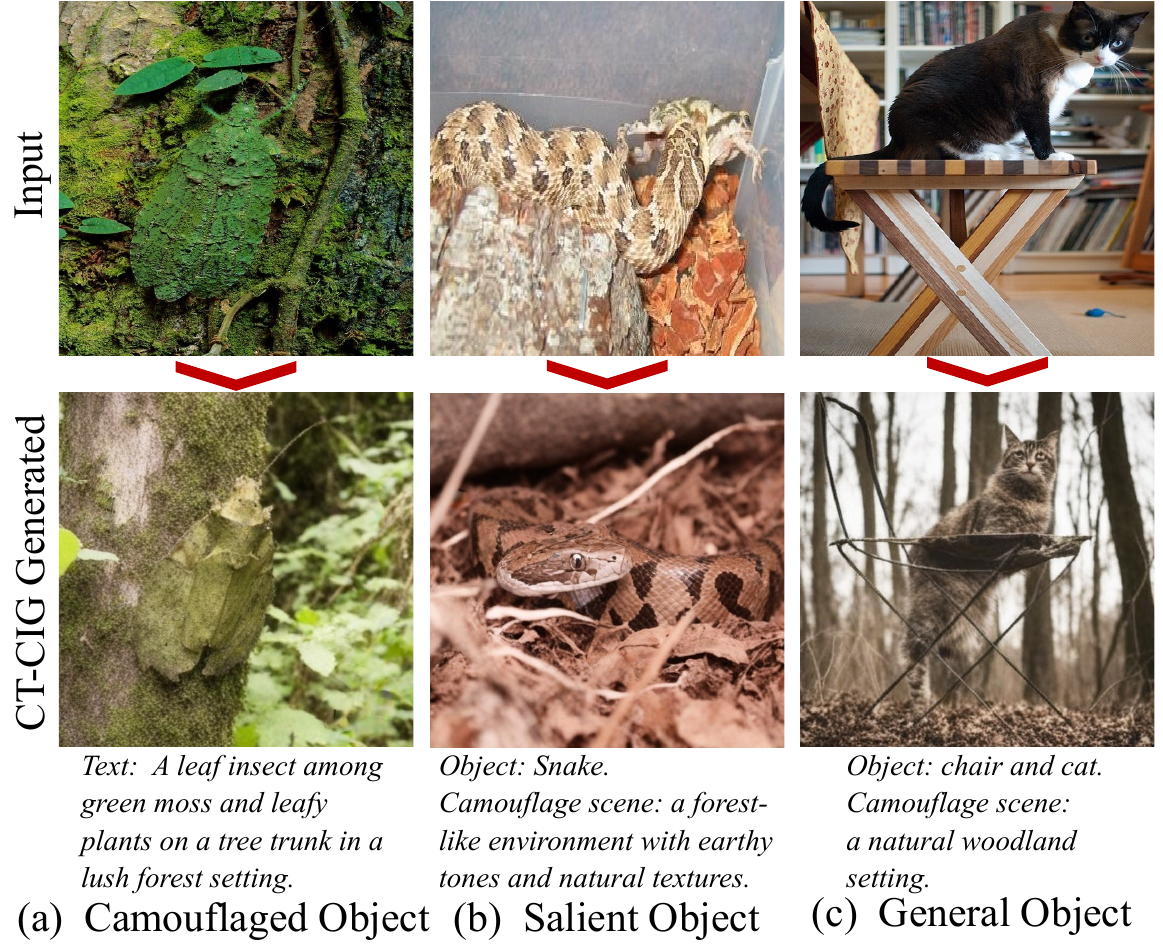}
    \caption{Example images generated by CT-CIG, which proves its ability to handle objects of different attributes.}
    \label{fig:examples}
    \vspace{-4mm}
\end{figure}

The technique of generating digital camouflage images can be dated back to \cite{chu2010camouflage} via hand-crafted feature processing. Current CIG methods generally fall into two paradigms. 
The first paradigm performs camouflage by altering the color and texture patterns of objects to blend themselves harmoniously into arbitrary backgrounds \cite{zhang2020deep, li2022location, gao2025ptdiffusion}, concluded as Background fitting. 
Despite the artistic beauty and exquisite visual concealment illusion, they fail to match the camouflage regularity that the natural world represents due to the damage to objects' appearances and ignorance of the logic between foreground objects and background images (e.g., a tiger face in a mountain).
The second paradigm, on the other hand, leverages generative models \cite{isola2017image, ho2020denoising, rombach2022high} to outpaint background surroundings with visual color consistency to foreground objects guided by their inherent features \cite{zhang2023camouflaged, zhao2024lake, das2025camouflage}, which can be viewed as Foreground guiding.
However, they fail to equip sufficient semantic consideration for the generation of backgrounds, which causes severe artifacts and makes the backgrounds unrealistic.

Motivated by the problems mentioned above, we aim to design a CIG method that can produce realistic and naturalistic camouflage images, and finally propose CT-CIG.
Examples of generated camouflage images are displayed in Fig. \ref{fig:examples}.
Specifically, CT-CIG employs stable diffusion to handle input images from existing COD datasets, with the corresponding text descriptions for the content guidance on both foreground objects and background environments.
Compensating for the absence of text prompts in COD datasets, we set a Camouflage-Revealing Dialogue Mechanism (CRDM) to exploit the image perception and contextual comprehension of Large Vision Language Models (VLM) \cite{li2023blip, wang2024qwen2} to obtain camouflage-sensed and semantically aligned text descriptions, which are of vital significance towards realistic camouflage generation.
Furthermore, a lightweight control network is utilized to handle binary masks of camouflaged objects and provide signals of objects' shapes and locations for the diffusion process.
To reduce pattern artifacts raised from noisy latents and endow the generated images with greater fidelity and more intricate details, we additionally design a Frequency Interaction Refinement Module (FIRM), which is capable of capturing high-frequency textures by attention weighing on Fast Fourier Transform (FFT) features \cite{campbell1968application}. Our key contributions can be summarized as follows:

\begin{itemize}
    \item We leverage VLMs and develop a Camouflage-Revealing Dialogue Mechanism to generate high-quality text prompts for camouflage images, which makes the paradigm of Text-guided Camouflage Images Generation implementable.
    \item We propose CT-CIG, a Controllable Text-guided Camouflage Images Generation method that leverages the powerful ability of diffusion to generate camouflage images with natural realism, along with a lightweight controller for geometric guidance of camouflaged objects and a Frequency Interaction Refinement Module to ensure the texture stability and content fidelity. 
    \item Experiments on the evaluation of metrics such as FID, KID, and CLIPScore, effectively demonstrate the fidelity of generated images and the descriptive accuracy of the generated text prompts. Related vision comparisons with previous methods validate CT-CIG's ability to produce photorealistic and logically feasible camouflage images.
\end{itemize}

\section{Related Works}

\subsection{Camouflage Images Generation}
The early CIG method \cite{chu2010camouflage} exploits hand-made features and devises a specialist algorithm for camouflage texture synthesis.
DCI \cite{zhang2020deep} and LCG-Net \cite{li2022location} leverage convolution networks \cite{simonyan2014very} on both foreground and background images and utilize feature fusion strategies inspired by style transfer \cite{huang2017arbitrary} to camouflage objects into specific backgrounds.
PTDiffusion \cite{gao2025ptdiffusion} transfers a reference image containing the target object to a text-instructed scene picture in which the object hides as a form of illusion.
These methods all follow the background-fitting paradigm without considering the plausibility of camouflaged objects and background surroundings, which makes camouflage a representation of vision art and optical illusion rather than a natural phenomenon.

Regarding foreground-guiding CIG methods, Generative Adversarial Networks (GAN) are used to synthesize artificial camouflage images based on the input of real images and masks from COD datasets \cite{zhang2023camouflaged, lamdouar2023making, he2024hestrategic}. 
LAKE-RED \cite{zhao2024lake} outpaints backgrounds that have similar colors to the foreground objects by knowledge retrieval with the combination of VQVAE \cite{van2017neural} and latent diffusion.
FACIG \cite{chen2025foreground} further refines the feature integration manner to reduce foreground distortion.
Due to the lack of any components to facilitate the background's semantic understanding, their generated images exhibit severe texture artifacts that fail to realize natural realism.
CamOT \cite{das2025camouflage} alleviates this problem to some degree by constructing a representation engineering work on clip-processed embeddings \cite{radford2021learning}.
However, ensuring logical feasibility and content controlability remains a challenge, since their prompts contain merely the literal word "background" and object labels in COD datasets.
Our CT-CIG, with VLM-produced prompts, opens the paradigm of Text-guided CIG, which enables generating logic-feasible and realistic camouflage images.

\subsection{Vision Language Models}

Large Vision Language Models (VLM) are capable of understanding the visual-language correlation for various tasks (e.g., object recognition, segmentation, graphic and image understanding) by learning rich vision (photo or video) text pairs that are almost infinite at web scale \cite{zhang2024vision}. 
Therefore, VLMs excel in zero-shot prediction on different tasks. \cite{li2025vlmsurvey}.
Pioneer VLMs such as CLIP \cite{radford2021learning} and BLIP \cite{li2023blip} are built by training the integrated vision encoder and the text encoder from scratch. In contrast, subsequent powerful VLMs like LLaVA \cite{liu2023llava} and Qwen-VL \cite{bai2023qwenvl} exploit LLMs as a backbone and design projectors for visual-language alignment in a shared embedding space.
Camobj-Llava \cite{ruan2025mm} is the first VLM that specializes in understanding camouflage scenarios by learning camouflage content produced by GPT-4o.
We choose Qwen2.5-VL \cite{bai2025qwen25vltechnicalreport} and impose Camouflage-Revealing Dialogue Mechanism to obtain text prompts paired with camouflage images for consideration of both performance and deployment convenience.

\subsection{Diffusion Models for Controllable Generation}

Diffusion \cite{ho2020denoising, song2020denoising} is widely used in various image generation tasks for its powerful ability to predict and eliminate noise.
LDM \cite{rombach2022high} transforms pixel images to latent features and performs diffusion in the latent space.
With support for extra condition inputs, such as text prompts, it succeeds in producing content-controlled outcomes and pioneers downstream tasks like layout-to-image \cite{wang2024instancediffusion}, image inpainting \cite{chen2024improving} or outpainting \cite{eshratifar2024salient}, and editing \cite{nguyen2025swiftedit}.
ControlNet \cite{zhang2023controlnet} is subsequently proposed to enable the generation of images with structural consistency to the multi-modal control signals (e.g., depth map, instance segmentation, scribble) that are handled by parallelism blocks in the stable diffusion UNet encoder, adaptively grasping the related spatial distributions.
ControlNext \cite{peng2024controlnext} furthermore processes external control signals only with a simple, lightweight network, which significantly accelerates training and serves as the foundation of CT-CIG.

\begin{figure*}[t]
\centering
\includegraphics[width=\textwidth]{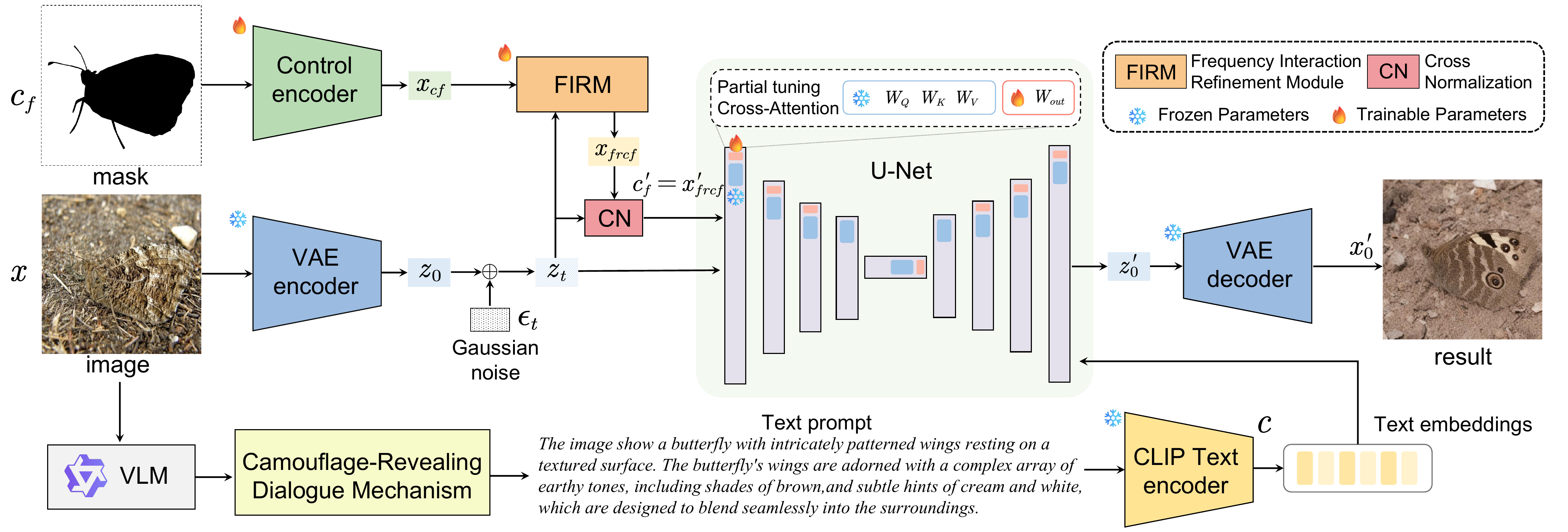} 
\caption{Overall framework of our proposed CT-CIG, which performs camouflage generation via three steps. (1) Extracting features of input images and masks through VAE and controller, followed by control augmentation through FIRM and CN. (2) Generating text prompts from the VLM through CRDM and using the CLIP encoder to obtain embeddings. (3) Performing controllable stable diffusion and generating results. }
\label{fig:architecture}
\vspace{-4mm}
\end{figure*}

\begin{figure*}
\centering
\includegraphics[width=\textwidth]{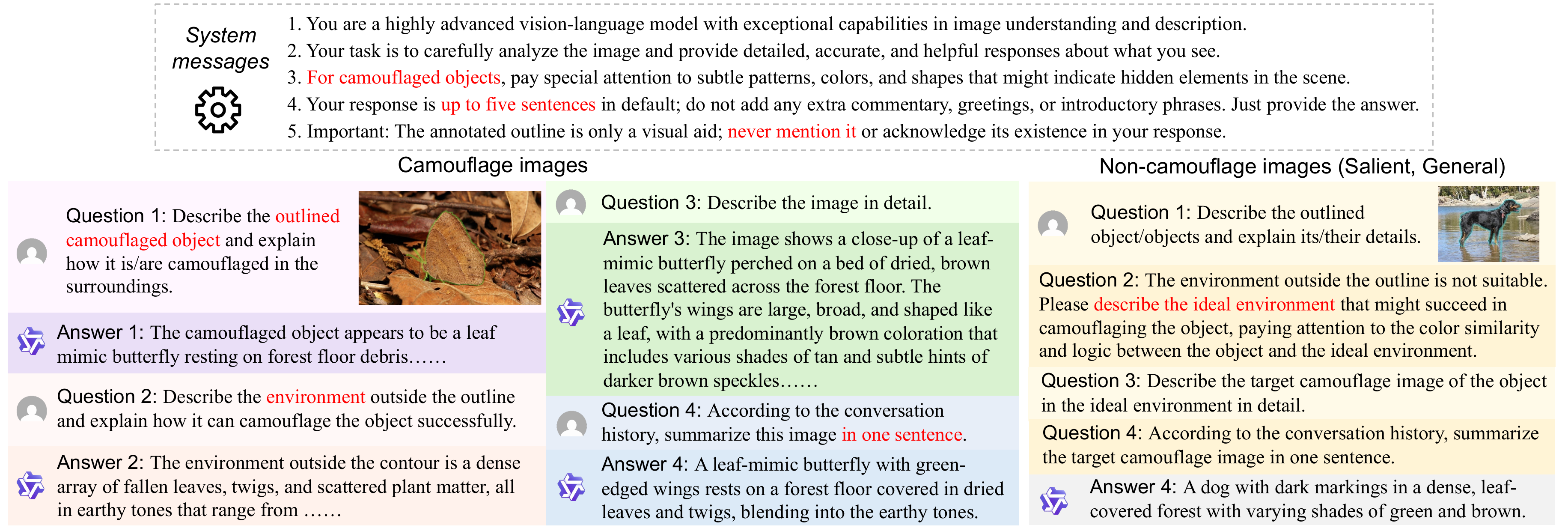}
\caption{Intra details of the Camouflage-Revealing Dialogue Mechanism. VLMs should obey the rules in system messages to produce answers that meet our requirements. Different queries are designed for camouflage images and non-camouflage images to guide them to generate camouflage-originated responses.}
\label{fig:mdm}
\vspace{-4mm}
\end{figure*}

\section{Methodology}
\subsection{Preliminaries}
The core of diffusion lies in predicting noise added to data under the paradigm of forward diffusion and reverse denoising.
DDPM \cite{ho2020denoising} estimates the posterior $p_{\theta}\left( x_{t-1}|x_t \right)$ with its corresponding noise adding process formulated as:
\begin{small}
 \begin{equation}
    x_t = \sqrt{\alpha_t}x_{t-1} + \sqrt{1-\alpha_{t}}\epsilon=\sqrt{\bar{\alpha_t}}x_0 + \sqrt{1-\bar{\alpha_{t}}}\epsilon,
    \label{eq1}
\end{equation}   
\end{small}where $x_0$ is the original image and $x_t$ is the noisy version at timesetp $t$, $\alpha_t$ denotes the diffusion constant and $\bar{\alpha_t}=\prod^{t}_{i=1}\alpha_{i}$.
DDIM overcomes DDPM's bottleneck of inference time raised from the Markov Chain to support step-skipped prediction by redefining $q(x_t|x_0) \sim \mathcal{N}(\sqrt{\alpha_t}x_0,(1-\alpha_t)I)$. 
The posterior also matches a normal distribution under this assumption from Bayes' theorem:
\begin{small}
  \begin{equation}
    q(x_{t-1}|x_t, x_0)=\frac{q(x_t|x_{t-1},x_0)q(x_{t-1}|x_0)}{q(x_t|x_0)}.
    \label{eq2}
\end{equation}  
\end{small}

It is evident that all elements on the right-hand side of Eq. \ref{eq2} can be reparameterized via Eq. \ref{eq1}. Therefore, the distribution of $ p_\theta(x_{t-1}|x_t)$ can be formulated as:
\begin{small}
  \begin{equation}
    \mathcal{N}(\sqrt{\alpha_{t-1}}\frac{x_t-\sqrt{1-\alpha_t}\epsilon_{\theta}}{\sqrt{\alpha_t}} +\sqrt{1-\alpha_{t-1}-\delta^{2}_{t}}\epsilon_{\theta},\delta^{2}_{t}I),
    \label{eq3}
\end{equation}  
\end{small}where $\delta^{2}_{t}$ can be valued as zero and $t-1$ can be replaced with previous timesteps to realize step-skipped sampling. More details can be found in \cite{song2020denoising}.
To reduce computation cost, LDM \cite{rombach2022high} utilizes a pre-trained autoencoder $\varepsilon$ \cite{kingma2013vae} to transform $x$ to a smaller latent representation $z_0=\varepsilon(x_0)$ and then performs diffusion. The training objective is to optimize noise prediction $\epsilon_\theta(z_t,t)$ via MSE loss:
\begin{small}
  \begin{equation}
    \mathcal{L}_{SD}=\mathrm{E}_{\varepsilon(x),t,\epsilon\sim\mathcal{N}(0,1)} [\left\|\epsilon_\theta(z_t,t)-\epsilon\right\|^{2}_{2}].
    \label{eq:loss_ldm}
\end{equation}  
\end{small}

Extra condition $c$ (e.g., text prompt) is optional in LDM for task-specific training, and ControlNeXt further integrates control signal $c_f$. The loss function can be calculated as:
\begin{small}
  \begin{equation}
    \mathcal{L}_{SD}=\mathrm{E}_{\varepsilon(x),t,c,c_f,\epsilon\sim\mathcal{N}(0,1)}[\left\|\epsilon_\theta(z_t,t,c,c_f)-\epsilon\right\|^{2}_{2}].
    \label{eq:loss_ldm_cond}
\end{equation}  
\end{small}

\subsection{Framework of CT-CIG}
To effectively synthesize camouflage images, CT-CIG must understand the intricate traits of natural camouflage, including object concealment, color consistency, texture similarity, and logical plausibility between objects and their background surroundings.
All traits except logical plausibility are learnable through training with authentic camouflage images.
Logical plausibility is not explicitly exhibited in the pixel domain but is presented implicitly in the semantic domain.
To compensate for the gap, text prompts paired with training images, which are available through VLMs and contain detailed semantic content, serve as conditional information for training CT-CIG.
Inspired by ControlNeXt \cite{peng2024controlnext}, binary masks of objects in COD datasets can be used more effectively as spatial guidance to control the location and shape of generated objects.

The general framework of CT-CIG is illustrated in Fig. \ref{fig:architecture}.
$x\in\mathcal{R}^{3\times h\times w}$ and $c_f\in\mathcal{R}^{1\times h \times w}$ are paired RGB and mask images.
Text prompt that describes $x$ is generated through Vision Question Answering (VQA) in VLM by the CRDM and subsequently fed into a CLIP Text-encoder to form text embeddings $c$.
We use a VAE to encode the RGB camouflage image to the latent $z_0\in\mathcal{R}^{4\times64\times64}$ and subsequently add Gaussian noise to obtain $z_t$, where $t$ is randomly sampled from $[0,T]$ ($T=1000$). 
A lightweight controller, comprises Resblocks \cite{he2016deep}, is exploited to encode $c_f$ to get the control feature $x_{cf}$.
FIRM then processes $x_{cf}$ along with $z_t$ to calculate its frequency-enhanced $x_{frcf}$ to grasp high-frequency texture details, followed by Cross Normalization (CN) to obtain the mean and variance aligned control feature $x^{\prime}_{frcf}$. 
After that, the denoising procedure to recover latent noise-free $z^{\prime}_{0}$ from $z_t$ with the condition of $c$ and $x^{\prime}_{frcf}$ is conducted via UNet \cite{ronneberger2015unet} based stable diffusion.
Finally, a VAE decoder transforms $z^{\prime}_{0}$ back to image domain and generates the result camouflage image $x^{\prime}_0$. Parameters in the controller, FIRM, and linear projectors in UNet's cross-attention blocks are trainable, while the remaining parts are kept frozen.

\subsubsection{Camouflage-Revealing Dialogue Mechanism.}
As shown in Fig. \ref{fig:mdm}, Camouflage-Revealing Dialogue Mechanism (CRDM) leverages VLM's ability in visual perception and contextual comprehension.
We design four questions for each image with camouflage-guided instructions that gradually teach the VLM to make camouflage-oriented descriptions. 
All images are pre-processed by using random semi-transparent colored outlines to annotate object boundaries, where they directly adjoin the backgrounds and the key insights of camouflage lie.
Thanks to the semi-transparent effect, it not only helps the VLM to localize and familiarize the camouflaged object but also preserves the related boundary pixel details to understand the camouflage paradigm.
For camouflage images, questions 1 and 2 are designed to obtain the descriptions of objects and the surrounding environments, along with their relationships.
Subsequently, these descriptions are reorganized to form detailed prompts through question 3.
The final question 4 is defined to review all contents and summarize each prompt into one sentence.
We denoted the detailed and summarized prompt as $T_{detail}$ and $T_{simple}$.
Despite a suitable dialogue mechanism to handle camouflage images, the descriptions of background environments act as negative knowledge regarding non-camouflage images (e.g., salient object images and general images).
Therefore, we ask the VLM to imagine an ideal scenery that might successfully camouflage the object in question 2 and produce related $T_{detail}$ and $T_{simple}$ in later questions, as illustrated in the right part of Fig. \ref{fig:mdm}.
Additionally, we pre-define some system messages to aid VLMs for camouflage perception and to force VLMs to avoid producing any unrelated and redundant content.
$T_{simple}$ is utilized during inference for generation diversity, while $T_{detail}$ is used for training CT-CIG because it encapsulates more intricate information, which forces CT-CIG to learn complex camouflage patterns and prevents catastrophic forgetting.
\vspace{-4mm}

\subsubsection{Frequency Interaction Refinement Module}
Compared to other controls such as depth map, infrared map, or object canny, binary masks only indicate rough position and geometric cues for target camouflaged objects, without subtle information about spatial hierarchy or objects' intra appearance.
Consequently, $x_{cf}$ encoded by the controller is information-deficient. Through easy control convergence, it has the potential risk of producing texture artifacts and unnatural hallucinations. 
To address this issue, we propose the Frequency Interaction Refinement Module (FIRM), based on the Fourier Transform, to enhance the information granularity of $x_{cf}$.
According to the theory of the Fourier Spectrum, low frequencies contribute to the image's overall structural information, while high frequencies contribute to the image's texture and intricate pattern information \cite{campbell1968application, zhou2023low_fourier}. 
Therefore, it is possible to equip $x_{cf}$ with detailed texture representations learned from the image latent $z_t$ through FIRM.

\begin{figure}[t]
\centering
\includegraphics[width=\columnwidth]{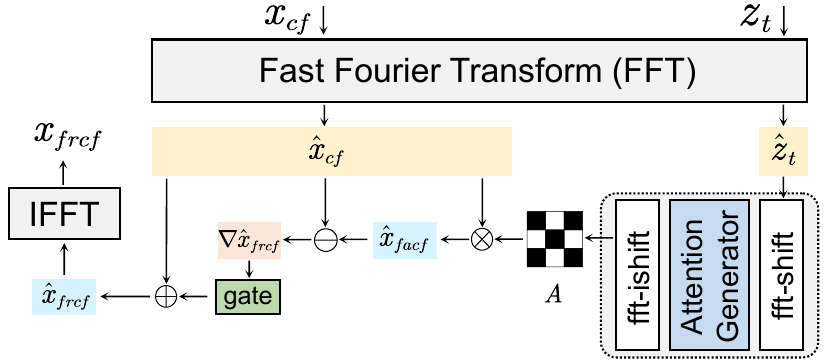} 
\caption{Frequency Interaction Refinement Module.}
\label{fig:firm}
\vspace{-4mm}
\end{figure}

Fig. \ref{fig:firm} displays the detailed pipeline of FIRM.
Designed to strengthen the granularity of the control signal guided by the input feature, $x_{cf}$ and $z_t$ are first applied to Fast Fourier Transform (FFT) to obtain related spectrums $\hat{x}_{cf}$ and $\hat{z}_t$ in the frequency domain.
An attention map $A$ that takes magnitude $|\hat{z}_t|$ as source data is generated by an Attention Generator block, which consists of 2 convolution layers. 
The initial pattern of $|\hat{z}_t|$ is not continuous because high-frequency components are separated in the center and corners. 
To fit convolution's local spatial perception, fftshift is employed to permutate a continuous spectrum where low frequencies are around the image center, with high frequencies uniformly arranged in the outer ring. 
The formulation for obtaining the frequency attention weights matrix $A$ is:
\begin{small}
 \begin{equation}
    A= \mathrm{ifft\text{-}shift}(\mathrm{AG}(\mathrm{fft\text{-}shift}(|\hat{z}_t|))),
    \label{eq:cal_A}
\end{equation}   
\end{small}where AG denotes Attention Generator, and ifft-shift restores the spectrum to its original pattern.
Subsequently, we interact $\hat{x}_{cf}$ with $A$ to capture subtle texture details and calculate the attention-enhanced control spectrum $\hat{x}_{facf}$ and the refinement gain $\nabla\hat{x}_{frcf}$.
The refinement gain is then added to $\hat{x}_{cf}$ with its intensity adaptively controlled by a learnable gate to form the frequently refined control spectrum $\hat{x}_{frcf}$. Related equations are presented as follows:
\begin{small}
\begin{align}
    \hat{x}_{facf} &= \hat{x}_{cf} \otimes A, \\
    \nabla{\hat{x}}_{frcf} &= \hat{x}_{facf} - \hat{x}_{cf}, \\
    \hat{x}_{frcf} &= \hat{x}_{cf} + gate \times \nabla{\hat{x}}_{frcf}.
\end{align}    
\end{small}

Finally, Inverse Fast Fourier Transform (IFFT) is applied to transfer the refined control back to the feature domain, denoted as $x_{frcf}$.
Compared to $x_{cf}$, this FIRM refined control feature is more fine-grained for ensuring robustness and facilitating the generation of complex camouflage textures.

\subsubsection{Diffusion with Cross Normalization} 
The discrepancy between the FIRM refined control feature and the noisy image latent poses a risk of color instability. Motivated by \cite{peng2024controlnext}, Cross Normalization (CN) is utilized as an alternative to the "zero convolution" layers in ControlNet to enhance training robustness. 
Initially, channel-wise mean and standard deviation of the control feature $x_{frcf}$ and latent $z_t$, notated as $\mu$ and $\sigma$, are calculated via:
\begin{small}
 \begin{equation}
    \mu_z, \mu_{cf} = \frac{1}{n_1}\sum^{n_1}_{i=1}z_{t,i} , \frac{1}{n_2}\sum^{n_2}_{i=1}x_{cf,i},
    \label{eq:cal_mu}
    \vspace{-4mm}
\end{equation}

\begin{equation}
    \sigma^{2}_z,\sigma^{2}_{cf} = \frac{1}{n_1}\sum^{n_1}_{i=1}(z_{t,i}-\mu_{z})^2, \frac{1}{n_2}\sum^{n_2}_{i=1}(x_{cf,i} - \mu_{cf})^2.
    \label{eq:cal_std}
\end{equation}   
\end{small}

We subsequently perform CN that undergos $x_{frcf}$ standardization and $z_t$ affine transformation formulated as: 

\begin{small}
 \begin{equation}
    x^{\prime}_{frcf} = \mu_{z} + \frac{x_{frcf} - \mu_{cf}}{\sqrt{\sigma^2_{cf}+ \varepsilon}}\times \sigma_{z},
    \label{eq:cn}
\end{equation}   
\end{small}where we define $c^{\prime}_{f}=x^{\prime}_{frcf}$ as the final control signal.
This concise and ingenious strategy renders $c^{\prime}_{f}$ distributally consistent with $z_t$, effectively facilitating training.

\subsubsection{Training optimization}
Apart from the condition integrated diffusion loss $\mathcal{L}_{SD}$, we empoly an LPIPS perceptual loss \cite{zhang2018unreasonable} to minimize the feature discrepancy of the predicted result $x^{\prime}_0$ and the input image $x_0$, which benefits the generation of natural and realistic images. The perceptual loss is calculated as:
\begin{small}
  \begin{equation}
    \mathcal{L}_{\mathrm{Lpips}} = \sum_{l}\frac{1}{h_lw_l}\sum_{h,w}\left\|\gamma_l \cdot (f^{\prime l}_{0} -f^{l}_{0}) \right\|^{2}_{2},
    \label{eq:lpips}
\end{equation}  
\end{small}where $f^{\prime}_0$, $f_0$ are VGG encoded feature stacks of $x^{\prime}_0$ and $x_0$, $l$ and $\gamma$ denote the stack id and scale factor. 
The overall loss function is formulated below, with perceptual loss multiplied by the corresponding weight.

\begin{small}
 \begin{equation}
    \mathcal{L}=\mathcal{L}_{SD} + \lambda_{\mathrm{Lpips}} \cdot \mathcal{L}_{\mathrm{Lpips}}.
    \label{eq:loss}
\end{equation}   
\end{small}

\begin{figure*}[t]
\centering
\includegraphics[width=0.95\textwidth]{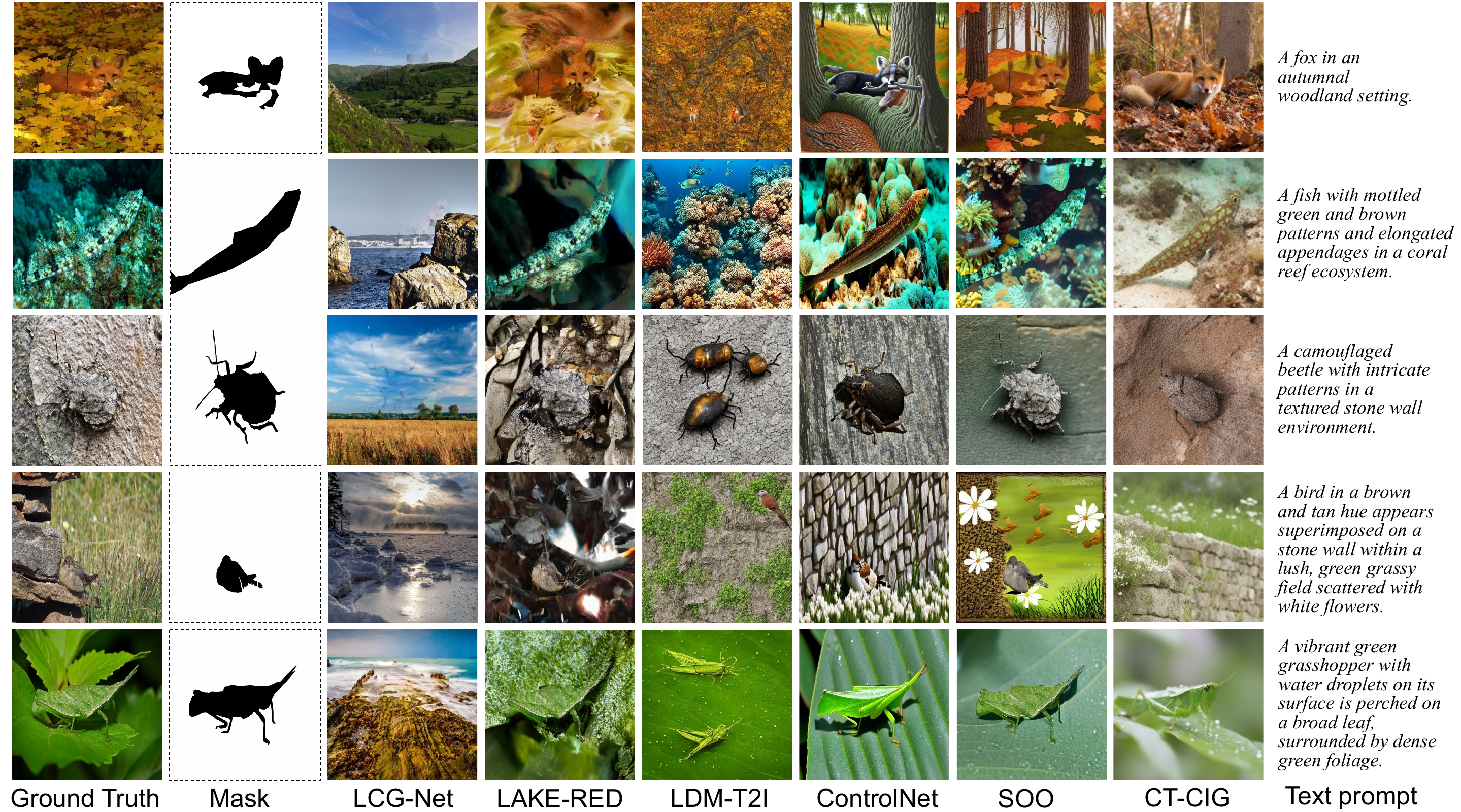}
\caption{Results of generated images with different methods. The first two columns show the real images and paired masks in the COD datasets. Backgrounds in column 3 are randomly selected. Methods in columns 5-8 require text prompts as a condition. All methods take masks as shape guidance except LDM-T2I.}
\label{fig:qual}
\vspace{-4mm}
\end{figure*}

\section{Experiments}

\subsection{Experimental Settings}
\subsubsection{Datasets and metrics}
We utilize the LAKE-RED dataset \cite{zhao2024lake}, which comprises 4040 images for training, 19419 images for validation, and 5066 images for evaluation.
The training dataset consists of camouflage images from COD10K\cite{fan2020camouflaged} and CAMO \cite{le2019CAMO}.
There exist three subsets in the test dataset, including Camouflaged Objects \cite{lv2021simultaneously}, Salient Objects \cite{wang2017duts}, and General Objects \cite{lin2014microsoft}; each category contains 6473 images.
We process these images through VLM to form related image-text pairs and use the CLIP Score \cite{hessel2021clipscore} to measure image-text alignment in both training and testing.
Following previous works, we choose FID \cite{binkowski2018fid} and KID \cite{heusel2017kid} to quantify the quality of generated images, taking the 5066 camouflage images from COD10K as the evaluation benchmark.

\subsubsection{Implementation Details}
To obtain text prompts, we build Qwen2.5-VL via the VLLM framework and perform VQA with 2 NVIDIA RTX 4090 GPUs.
Subsequently, we implement CT-CIG based on ControlNext and take a pre-trained SDXL as the foundation of diffusion. During training, images and control masks are resized to $512\times512$ and then transformed to $128\times128$ latents with a batch size of 4.
We set the control scale as 1.2 and $\lambda_{LPIPS}$=1e-3.
For quick convergence, the learning rate of the lightweight controlnet and FIRM are set to 1e-4, which facilitates control signal adaptation, while SDXL UNet's learning rate is set to 5e-6 for cautious finetuning.
It takes around 8 hours to train our CT-CIG for 80 epochs with 4 NVIDIA RTX A5000 GPUs.

\subsection{VLM Choice}
VLMs vary in their ability to understand the input content and make responses.
We employ CRDM to several open-source VLMs \cite{li2023blip, liu2023llava, team2025gemma, bai2025qwen25vltechnicalreport} and use CLIPScore to measure the quality of their produced content. 
As shown in Table \ref{table1}, $T_{detail}$ is likely to outperform $T_{simple}$ under the same VLM because it contains more corpus that explains images at a finer grain.
Qwen2.5-VL is chosen as our VLM backbone for its superior text-image alignment performance.
\begin{table}
\centering
{\fontsize{9pt}{1pt}\selectfont

\begin{tblr}{
  cells={c},
  hline{1,6} = {-}{1pt},
  hline{2} = {-}{},
}
\textbf{Method}     & \textbf{CLIP simple $\uparrow$} & \textbf{CLIP detail $\uparrow$} \\
BLIP2-2.7B      &  0.2461          & 0.2859           \\
LLaVA-13B      & 0.2986           & 0.2969           \\
Gemma3-4B      & 0.3127        & 0.3136           \\
Qwen2.5-VL-7B & \textbf{0.3183}           & \textbf{0.3242}           
\end{tblr}
}
\caption{CLIPScores of detailed and summarized prompts generated by employing CDRM to different VLMs.}
\label{table1}
\vspace{-4mm}
\end{table}

\subsection{Comparison with SOTA methods}
We compare our CT-CIG with 11 SOTA methods; some are specialized in CIG \cite{zhang2020deep, li2022location, zhao2024lake}, while others can be adapted to camouflage generation \cite{perez2003poisson,huang2017arbitrary,zheng2022TFill, rombach2022high, lugmayr2023RePaint-L, zhang2023controlnet, eshratifar2024salient}, covering the paradigms of Background fitting, Foreground guiding, and Text guiding.

\subsubsection{Qualitative Results}
Fig. \ref{fig:qual} displays the generated images of different methods. 
The background-fitting LCGNet camouflages objects into arbitrary background images excessively, which causes severe distortion and makes them barely visible.
LAKERED outpaints the background regions with similar colors guided by the foreground objects.
Both of these paradigms ignore the semantic relationship between camouflaged objects and environments, which contributes to little logical plausibility.
With the integration of text prompts, LDM-T2I produces images with overall semantic tendency, but their compositions are disorganized without control guidance.
Taking masks as a control signal, ControlNet successfully generates objects in the correct location and shapes. However, the generative artifacts lead to a lack of photorealism, which places them in the uncanny valley.
Salient Object Outpainting (SOO) is designed to fill the background with general prompts, which can't guarantee its fitness for camouflage.
Our CT-CIG, in contrast, is capable of generating photorealistic camouflage images with detailed textures, depth of field, and contextual coherence.

\begin{table*}[]
\centering
{\fontsize{9pt}{1pt}\selectfont
\begin{tblr}{
  row{even} = {c},
  row{1} = {c},
  row{3} = {c},
  row{5} = {c},
  row{7} = {c},
  row{9} = {c},
  row{11} = {c},
  row{13} = {c},
  cell{1}{1} = {r=2}{},
  cell{1}{2} = {r=2}{},
  cell{1}{3} = {c=2}{},
  cell{1}{5} = {c=2}{},
  cell{1}{7} = {c=2}{},
  cell{1}{9} = {c=3}{},
  cell{3}{1} = {r=4}{},
  cell{7}{1} = {r=4}{},
  cell{11}{1} = {r=4}{},
  cell{11}{2} = {c},
  cell{11}{3} = {c},
  cell{11}{4} = {c},
  cell{11}{5} = {c},
  cell{11}{6} = {c},
  cell{11}{7} = {c},
  cell{11}{8} = {c},
  cell{11}{9} = {c},
  cell{11}{10} = {c},
  cell{11}{11} = {c},
  hline{1,15} = {-}{1.2pt},
  hline{3,7,11} = {-}{},
  hline{2} = {3-11}{},
}
\textbf{Paradigm}                            & \textbf{Method}      & \textbf{Camouflaged Objects} &        & \textbf{Salient Objects} &        & \textbf{General Objects} &        & \textbf{Overall} &        &        \\
                                    &             & FID $\downarrow$                 & KID $\downarrow$   & FID $\downarrow$            & KID$\downarrow$    & FID $\downarrow$            & KID$\downarrow$    & FID $\downarrow$    & KID $\downarrow$   & CLIP $\uparrow$  \\
{Background\\fitting    }           & AB \shortcite{perez2003poisson}          & 117.11              & 0.0645 & 126.78          & 0.0614 & 133.89          & 0.0645 & 120.21  & 0.0623 & -      \\
                                    & AdAIN \shortcite{huang2017arbitrary}      & 125.16              & 0.0721 & 133.20          & 0.0702 & 136.93          & 0.0714 & 126.94  & 0.0703 & -      \\
                                    & DCI \shortcite{zhang2020deep}        & 130.21              & 0.0689 & 134.92          & 0.0665 & 137.99          & 0.0690 & 130.52  & 0.0673 & -      \\
                                    & LCGNet \shortcite{li2022location}     & 129.80              & 0.0504 & 136.24          & 0.0597 & 132.64          & 0.0548 & 129.88  & 0.0550 & -      \\
{Foreground\\guiding              } & TFill \shortcite{zheng2022TFill}     & 63.74               & 0.0336 & 96.91           & 0.0453 & 122.44          & 0.0747 & 80.39   & 0.0438 & -      \\
                                    & LDM-Inpaint \shortcite{rombach2022high} & 58.65               & 0.0380 & 107.38          & 0.0524 & 129.04          & 0.0748 & 84.48   & 0.0488 & -      \\
                                    & RePaint-L \shortcite{lugmayr2023RePaint-L}  & 76.80               & 0.0459 & 114.96          & 0.0497 & 136.18          & 0.0686 & 96.14   & 0.0498 & -      \\
                                    & LAKERED \shortcite{zhao2024lake}    & 39.55               & 0.0212 & 88.70           & 0.0428 & \textbf{102.67}          & 0.0625 & 64.27   & 0.0355 & -      \\
{Text \\ guiding}                         & LDM-T2I \shortcite{rombach2022high}    & 51.24   & 0.0206      & 102.04               & \textbf{0.0217}      & 120.41  & \underline{0.0304}     & 73.51       & 0.0261      & 0.2873      \\
                                    & ControlNet \shortcite{zhang2023controlnet}  & 39.67               & 0.0121 & \underline{81.72}           & 0.0303 & \underline{102.94}          & 0.0422 & \underline{59.52}   & 0.0227 & 0.2950     \\
                                    & SOO \shortcite{eshratifar2024salient}        & \underline{30.92}               & \textbf{0.0056} & 89.46           & 0.0267 & 117.31          & 0.0423 & 59.75   & \underline{0.0187} & \underline{0.3043} \\
                                    & CT-CIG (ours) & \textbf{30.59}               & \underline{0.0085} & \textbf{81.60}           & \underline{0.0230} & 104.46          & \textbf{0.0241} & \textbf{52.88}   & \textbf{0.0169} & \textbf{0.3243} 
\end{tblr}
}
\caption{Quantitative evaluation results on FID, KID, and CLIPScore for generated images of CT-CIG compared with 11 SOTA methods. The best results are highlighted in bold, and the second-best results are underlined. }
\label{table2}
\vspace{-3mm}
\end{table*}

\begin{figure*}[t]
    \centering
    \includegraphics[width=0.96\textwidth]{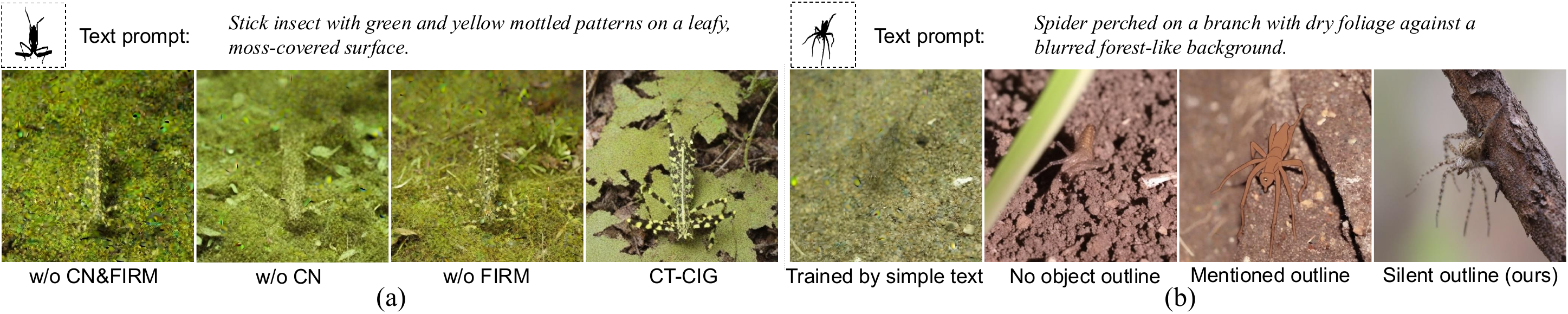}
    \caption{Visionalizations of ablation studies, including (a) the role of FIRM and CN in CT-CIG, and (b) training with text prompts that were generated via different query settings in CRDM.}
    \label{fig:ablation}
    \vspace{-4mm}
\end{figure*}

\begin{table}
\centering
{\fontsize{9pt}{1pt}\selectfont
\begin{tblr}{
  cells = {c},
  hline{1,6} = {-}{1pt},
  hline{2} = {-}{},
}
\textbf{Configration} & \textbf{FID $\downarrow$} & \textbf{KID $\downarrow$}\\
w/o FIRM\&CN   &   32.37  &   \textbf{0.0079}     \\
w/o CN       &   33.99  &   0.0114  \\
w/o FIRM     &  31.66   &    0.0080 \\
CT-CIG        &  \textbf{30.59}   &   0.0085
\end{tblr}
}
\caption{Image quality quantification for ablating FIRM and Cross Nomalization. "w/o" denotes without.}
\label{table3}
\vspace{-6mm}
\end{table}

\vspace{-1mm}
\subsubsection{Quantitative Results}
According to the LAKERED test dataset, we split the generated images into three groups. 
Apart from FID and KID, we use CLIPScore in text-guided methods to evaluate the semantic alignment of generated images and text prompts.
Related results are shown in Table \ref{table2}.
Background fitting methods perform the worst in all object types due to the object appearance destruction.
Foreground guiding methods achieve some improvement by maintaining object consistency, but the textual artifacts in background regions limit their performance upperbound.
Large performance enhancement can be realized with the combination of mask control signal in text guiding methods.
Our method achieves the best performance on the image quality and text alignment in the overall setting, and remains highly competitive in category-specific entries. 
Holistically, CT-CIG presents the most compelling performance profile.

\subsection{Ablation Study}
As illustrated in Fig. \ref{fig:ablation} (a), the baseline model (without FIRM and CN) struggles to produce high-quality camouflage images. 
Adding Cross Normalization effectively reduces spurious color speckles.
The integration of FIRM further mitigates the issue of missing high-frequency details and enables CT-CIG to generate images with clear structures and subtle textures.
Results in Table \ref{table3} show the performance improvements upon incorporating FIRM and CN.
Fig. \ref{fig:ablation} (b) and Table \ref{table4} show the effect of different text prompt configurations in CRDM.
Training with $T_{simple}$ causes CT-CIG to suffer from catastrophic forgetting, producing results that are camouflaged but blurry.
We find that prompts generated without object outlines, even those achieving a high CLIPScore with the source image, often fail to match shape guidance.
When the restrictions on outline-related content are removed, the resulting prompts tend to include explicit descriptions of outlines, which can mislead the generation process and produce line-drawing-like artifacts. 
These ablation studies effectively demonstrate the necessity and effectiveness of each designed component in CT-CIG.

\begin{table}
\centering
{\fontsize{9pt}{1pt}\selectfont
\begin{tblr}{
  cells = {c},
  hline{1,6} = {-}{1pt},
  hline{2} = {-}{},
}
\textbf{Configration}                   & \textbf{CLIP Score $\uparrow$} & \textbf{FID $\downarrow$} & \textbf{KID $\downarrow$} \\
Trained by simple text & 0.3183         & 54.92   & 0.0387   \\
No object outline & \textbf{0.3247}        & 39.24   & 0.0112   \\
Mentioned outline   & 0.3218         & 39.79   & 0.0138   \\
Silent outline (ours)               & 0.3242          & \textbf{30.59}   & \textbf{0.0085} 
\end{tblr}
}
\caption{The impact of different text prompts obtaining configurations on model performance. "Silent outline" means anything directly related to object outlines is forbidden in text prompts.}
\label{table4}
\vspace{-6mm}
\end{table}

\section{Conclusion}
In this paper, we propose a Controllable Text-guided Camouflage Images Generation method, termed CT-CIG, which leverages a Stable Diffusion backbone with a lightweight controller to generate camouflage images with logical plausibility and natural realism. 
CT-CIG incorporates three key components: a Camouflage-Revealing Dialogue Mechanism that obtains text prompts with the aid of VLMs, a Frequency Interaction Refinement Module for capturing high-frequency texture details, and Cross Normalization to ensure generation stability. 
Rigorous experiments demonstrate the superior performance of CT-CIG.
We hope our work lays the groundwork for the Text-guided Camouflage Images Generation paradigm and will spur further research in this field.

\section*{Acknowledgements} 
This work was partially supported by the National Natural Science Foundation of China (No. U25A20533, No. 62276129), the Natural Science Foundation of Jiangsu Province (No. BK20250082), the Fundamental Research Funds for the Central Universities (No. NE2025010, No.NS2025038), and the Jiangsu Funding Program for Excellent Postdoctoral Talent (No. 2025ZB306).

\bibliography{Main_paper}

\newpage
\twocolumn[
  \begin{center}
    \vspace{4em}
    {\LARGE\bfseries Text-guided Controllable Diffusion for Realistic Camouflage Images Generation \par}
    \vspace{2em} 
    {\Large\bfseries Supplementary Material \par}
    \vspace{2em}   
  \end{center}
]

\section{A. Model Efficiency}
We provide explanations about the efficiency of our CT-CIG compared to other methods mentioned in the main paper, including the analysis on trainable parameters, inference speed, and memory cost, to prove the overall deployment convenience of CT-CIG.

\subsection{A.1 Parameters}
Components that require training in our CT-CIG include the ResNet-based lightweight controller, the Frequency Interaction Refinement Model (FIRM), and the linear projectors after cross-attention in the Stable Diffusion XL-based UNet \cite{podell2023sdxl2}.
Table \ref{tab:params_CTCIG} shows the scale of trainable and overall parameters of CT-CIG.
We succeeded in making the diffusion UNet adapted to camouflage scenes by finetuning only about 4\% of its total parameters, with an extra lightweight controller for geometric guidance and the FIRM for capturing texture details.
This trait of a few trainable parameters makes it possible to train CT-CIG on devices with limited computational resources.

\begin{table}[h]
\centering
 {
\fontsize{9pt}{1.2pt}\selectfont
\begin{tblr}{
  cells = {c},
  hline{1,8} = {-}{1.2pt},
  hline{2,7} = {-}{},
}
\textbf{Compontent}   & \textbf{Trainable params} & \textbf{Overall params} \\
VAE          & 0                & 83.65M         \\
Text encoder & 0                & 817.72M        \\
Controller   & 3.53M            & 3.53M          \\
FIRM         & 0.21M            & 0.21M          \\
UNet         & 102.48M          & 2567.46M       \\
\textbf{Total}        & \textbf{106.22M}          & \textbf{3472.57M}       
\end{tblr}
}   
\caption{Number of trainable and overall parameters in each component of CT-CIG, where 1M = $10\times6$.}
\label{tab:params_CTCIG}
\end{table}

Table \ref{tab:params_compare} shows the parameter scales of the compared diffusion-based methods mentioned in the main paper.
These methods vary in the overall number of parameters, which arises from their different choices in VAEs, Text encoders, and base UNet models. For instance, we choose SD1.5 in LDM-T2I, SD2.1 in ControlNet, SDXL in our CT-CIG, and customized models provided by the authors in LAKERED.
To enable the generation of camouflage images, all other methods require training the entire or a part of Unet-like blocks, which contributes a considerable number of trainable parameters.
Despite the enormous scale of overall parameters, our CT-CIG performs camouflage adaptation by only finetuning the final linear project layers of the cross-attention in SDXL, which effectively reduces training burden and achieves competitive results even under the settings of half-precision targeted at memory saving.

\begin{table}[h]
\centering
{\fontsize{9pt}{1.2pt} \selectfont
\begin{tblr}{
  cells = {c},
  hline{1,6} = {-}{1.2pt},
  hline{2} = {-}{},
}
\textbf{Method}     & \textbf{Trainable params} & \textbf{Overall params} & \textbf{Type} \\
LAKERED    & 387.25M          & 440.47M        & fp32   \\
LDM-T2I    & 1454.29M         & 1537.95M       & fp32   \\
ControlNet & 1,230,14M        & 1,667,83M      & fp32   \\
CT-CIG     & 106.22M          & 3472.57M       & fp16   
\end{tblr}
}
\caption{Number of trainable and overall parameters of CT-CIG compared with other diffusion-based methods.}
\label{tab:params_compare}
\end{table}

\subsection{A.2 Inference speed}
Another dimension to explain efficiency is to evaluate the speed of generating camouflage images.
This study is performed by using different methods to generate a tiny subset of images from the validation dataset, where we select 100 mask-prompt pairs for test and 10 extra pairs for GPU warmup that minimize error.
Once the testing process is finished, we record the total inference time and subsequently calculate the average inference time (Latency), Frames Per Second (FPS), and DDIM Steps per Second (SPS), as shown in Table \ref{tab:speed}.
The value of DDIM steps influences the Latency and FPS, while SPS directly measures the reversed image denoising speed.
It turns out that our CT-CIG achieves superior performance in terms of inference efficiency.

\begin{table}[h]
\centering
{\fontsize{9pt}{1.2pt}\selectfont
\begin{tblr}{
  cells = {c},
  hline{1,7} = {-}{1.2pt},
  hline{2} = {-}{},
}
\textbf{Method} & \textbf{DDIM steps} & \textbf{Latency $\downarrow$} & \textbf{FPS $\uparrow$} & \textbf{SPS $\uparrow$} \\
LAKERED         & 50                  & 4.989            & 0.20         & \textbf{10.02}        \\
LDM-T2I         & 50                  & 8.901            & 0.11         & 5.62         \\
ControlNet      & 20                  & 7.735            & 0.13         & 2.59         \\
SOO             & 20                  & 4.061            & 0.25         & 4.92         \\
CT-CIG (ours)   & 20                  & \textbf{2.085}            & \textbf{0.48}         & 9.59         
\end{tblr}
\caption{Results of metrics that reveal the inference speed. The units of Latency, FPS, and SPS are second/image, image/second, and steps/second. The best results are highlighted in bold.}
\label{tab:speed}
}
\end{table}

\begin{figure}[h]
    \centering
    \includegraphics[width=0.8\linewidth]{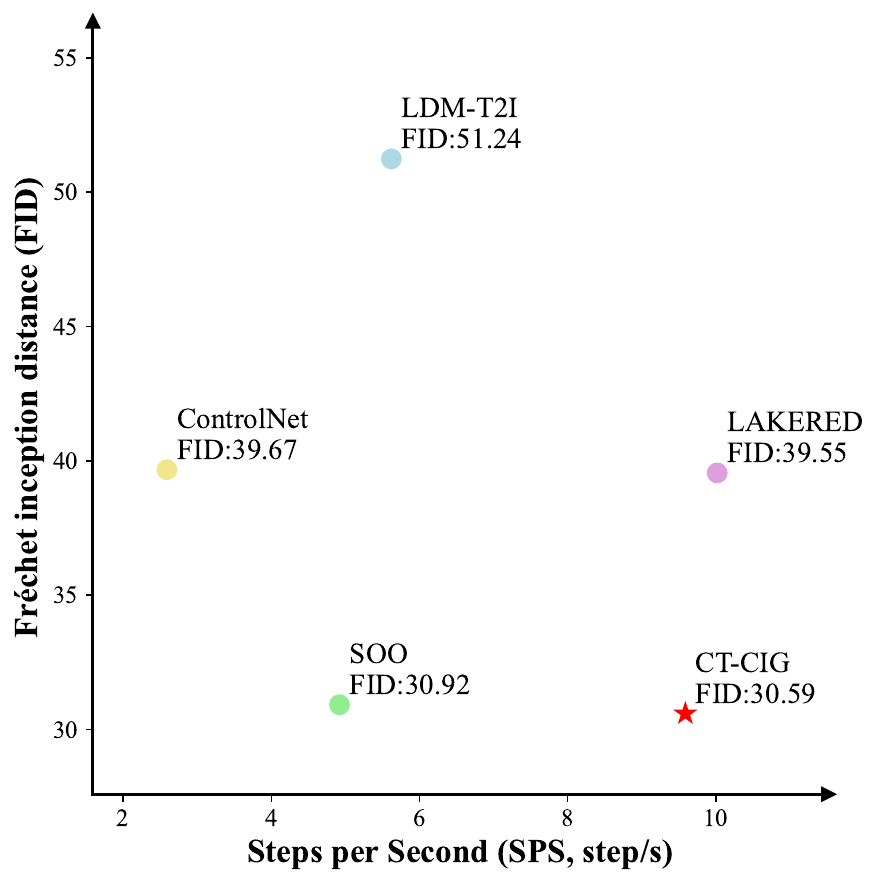}
    \caption{Scatter plot of different methods in terms of SPS-FID. Methods that appear in the lower right area have better performance.}
    \label{fig:scatter}
\end{figure}
We measure the Fréchet inception distance (FID) and SPS with the results visualized in Fig. \ref{fig:scatter} for the consideration of both model performance and inference efficiency.
Obviously, CT-CIG achieves the best trade-off between high image quality and fast inference speed.

\subsection{A.3 Memory cost}
In Table \ref{tab:GPUcost}, we provide the single RTX A5000 GPU memory usage results of different methods for generating camouflage images during inference.
Evidently, CT-CIG has remarkable strength in memory efficiency.
Compared to LDM-T2I, CT-CIG supports precise control of object shape with less GPU cost, and compared to LAKE-RED, CT-CIG enables the generation of natural camouflage patterns with plausible foreground-background logic via only a little memory sacrifice.

\begin{table}
\centering
{\fontsize{9pt}{1.2pt}\selectfont
\begin{tblr}{
  cells = {c},
  hline{1,7} = {-}{1.2pt},
  hline{2} = {-}{},
}
\textbf{Method} & \textbf{GPU Memory Cost (GB)} \\
LAKERED         & \textbf{8.46}                           \\
LDM-T2I         & 10.65                                   \\
ControlNet      & 16.89                                   \\
SOO             & 9.54                                    \\
CT-CIG(ours)    & \underline{8.74}                                    
\end{tblr}
\caption{Statistics on inference GPU memory usage of different methods. Top2 results are \textbf{bolded} and \underline{underlined}.}
\label{tab:GPUcost}
}
\vspace{-2mm}
\end{table}

\section{B. More analysis of the Ablation Study}
In this section, we provide more detailed information about the components mentioned in the ablation study to further substantiate our findings, including the impact of text prompts, FIRM's FFT size and gate initialization,  and the clever aspect of FIRM.
\subsection{B.1 Impact of text prompts on generated images.}

\begin{figure}[h]
    \centering
    \includegraphics[width=\linewidth]{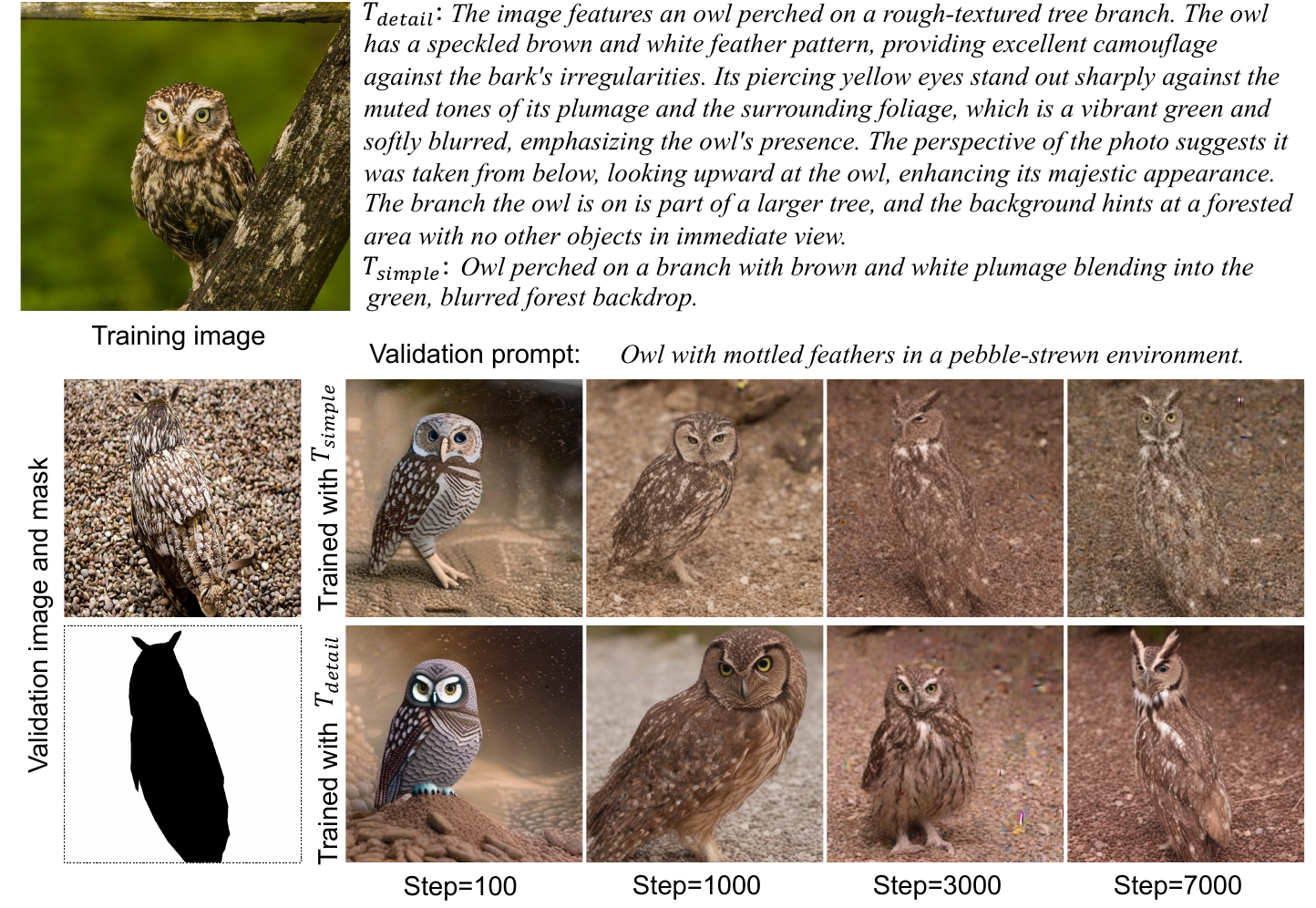}
    \caption{Visualization of the training progress using simple and detailed text prompts.}
    \label{fig:text1}
    \vspace{-2mm}
\end{figure}

We argue that training CT-CIG with detailed text prompts prevents the model from catastrophic forgetting \cite{kemker2018measuring2}. Fig. \ref{fig:text1} shows the validation results in the training progress with simple and detailed prompts.
There are no camouflage traits in the original steps because the alternation of parameters in the SDXL is insufficient to take effect.
Subsequently, the validated images exhibit camouflage but do not follow the shape control guidance.
When trained with $T_{simple}$, the semantic information is coarse. As the model adapts to the shape guidance, it identifies a simpler optimization path: generating a rudimentary object that broadly satisfies the prompt within the specified region is more efficient than rendering intricate details. Consequently, to expedite convergence on the control signal, the model progressively sacrifices its ability to generate fine-grained textures, a manifestation of catastrophic forgetting. While this process accelerates the convergence rate, it results in images that are perceptibly blurry and lack high-frequency details.
In contrast, training with $T_{detail}$ provides semantically rich and specific guidance. This enriched textual information acts as a regularizer, compelling the model to preserve its capacity for rendering complex details while simultaneously aligning with the shape control. As a result, the model is guided toward a more optimal solution that satisfies both the shape constraints and the detailed textual descriptions. This prevents the degradation of its generative capabilities, leading to a significant improvement in final image quality and fidelity.

\begin{figure}[t]
    \centering
    \includegraphics[width=\linewidth]{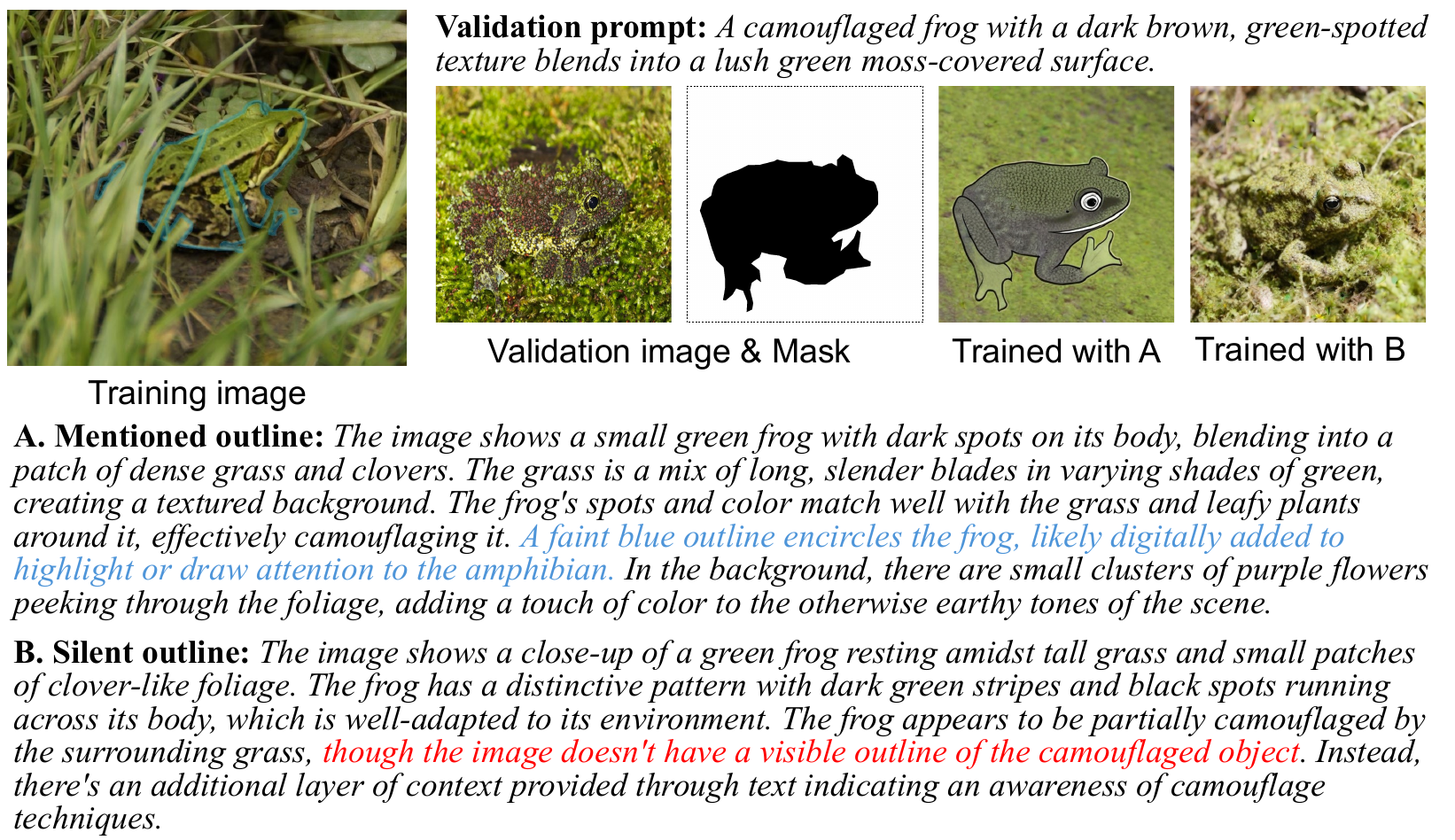}
    \caption{Visualization on the negative impact of explicit content of object outlines in text prompts. Config A denotes that there are no restrictions in the system messages of CRDM on image outlines, while config B denotes that any content related to the annotated outline is forbidden to be included in the generated texts.}
    \label{fig:text2}
    \vspace{-2mm}
\end{figure}

A further consideration in our text generation pipeline is the handling of object-annotated outlines. When generating text prompts for our training dataset using Qwen2.5-VL, we first annotated the objects in all images with semi-transparent colored outlines. This step was intended to make the camouflaged subjects more salient for the VLM. A critical challenge, however, is that any textual reference to these non-diegetic outlines could negatively bias the training of CT-CIG.
To mitigate this potential issue, we configured the system messages in our Camouflage-Revealing Dialogue Mechanism (CRDM) to explicitly forbid the VLM from mentioning any annotated outlines in its responses. The effectiveness of this approach is illustrated in Fig. \ref{fig:text2}, which contrasts the visual outcomes when the model is trained with prompts generated in the presence and absence of this restriction.

When the restriction concerning outlines is lifted, the text prompts generated by Qwen2.5-VL become populated with extraneous information about these outlines, as exemplified in prompt A. Such prompts mislead the model into an erroneous assumption that the target object should possess a prominent, visible contour. Consequently, the model is biased towards generating outputs in a line-art style, featuring objects with distinct outlines.
In contrast, with the restriction in place, the generated prompts actively avoid any reference to outlines or include descriptions that implicitly suppress their formation, as shown in prompt B. This ensures that the CT-CIG training process is not affected by these visual artifacts, enabling the generation of high-quality camouflage images.

\subsection{B.2 Impact of FFT size and gate in FIRM}
In the Frequency Interaction Refinement Module (FIRM), we transfer the control feature and the latent noise to the frequency domain via FFT, denoted as $\hat{x}_{cf}$ and $\hat{z}_t$, for the purpose of capturing high-frequency signals that reveal complex and subtle patterns of camouflage texture.
Generally, we keep the FFT transformed $\hat{x}_{cf}$ and $\hat{z}_t$ to the same size as their spatial features.
The gate is initialized to 0 and undergoes value adaptation during training, which ensures control signals are not affected by unstable frequency signals in the beginning and subsequently enhanced by optimized attention-adjusted frequency features.
We investigate the impact of FFT size and gate by altering their initialized values. Related quantitative benchmarks of generated images from the COD category in the LAKE-RED validation dataset are displayed in Table \ref{tab:FFT_gate}.
Both changing the initial value of the gate and the FFT scale factor lead to significant performance degradation.
Larger FFT size brings no information gain via padding but increases the computation burden and the risk of overfitting, while a smaller FFT size easily causes spectral aliasing.
\begin{table}[h]
\centering
{\fontsize{9pt}{1.2pt}\selectfont
\begin{tblr}{
  cells = {c},
  hline{1,6} = {-}{1.2pt},
  hline{2} = {-}{}
}
\textbf{Configuration}       & \textbf{FID}$\downarrow$   & \textbf{KID}$\downarrow$    \\
FFT $\times$ 2               & 45.76 & 0.0126 \\
FFT $\times$ 0.5             & 40.76 & 0.0157 \\
gate init=1         & 43.59 & 0.0168 \\
FFT $\times$ 1 + gate init=0 (ours) & \textbf{30.59} & \textbf{0.0085} 
\end{tblr}
}
\caption{Quantitative results on generating images with different settings of FFT size and gate initialization in FIRM. The best results are \textbf{bolded}.}
\label{tab:FFT_gate}
\vspace{-2mm}
\end{table}

\subsection{B.3 How FIRM captures texture details}
The capability of the Frequency Interaction Refine Module (FIRM) to capture high-frequency texture details, which was validated in our main paper's Ablation Study, warrants a deeper explanation. The ingenuity of FIRM lies in its architectural position, operating between the control features $x_{cf}$ and the noisy latent $z_t$.
The Stable Diffusion training regimen exposes FIRM to a dynamic spectrum of conditions determined by the random timestep $t$, This functionality is twofold:

1. Low-Noise Conditions (small $t$): The noisy latent $z_t$ is rich with information from the source image. FIRM capitalizes on this by directly modeling the fine-grained textural patterns present in $z_t$. 

2. High-Noise Conditions (large $t$): The latent $z_t$ is feature-depleted. In this case, FIRM relies on its previously acquired understanding from low-noise iterations to effectively process and model the highly noisy latent.

Random sampling of $t$ ensures that FIRM is consistently trained to handle information-rich and information-scarce latents. This endows the module with a robust capacity for texture detail acquisition. As a result, at inference time, when starting the denoising process from pure noise, FIRM can adeptly reconstruct and generate intricate, high-frequency textures.

\section{C. More qualitative results}
\begin{figure*}[t]
    \centering
    \includegraphics[width=0.9\linewidth]{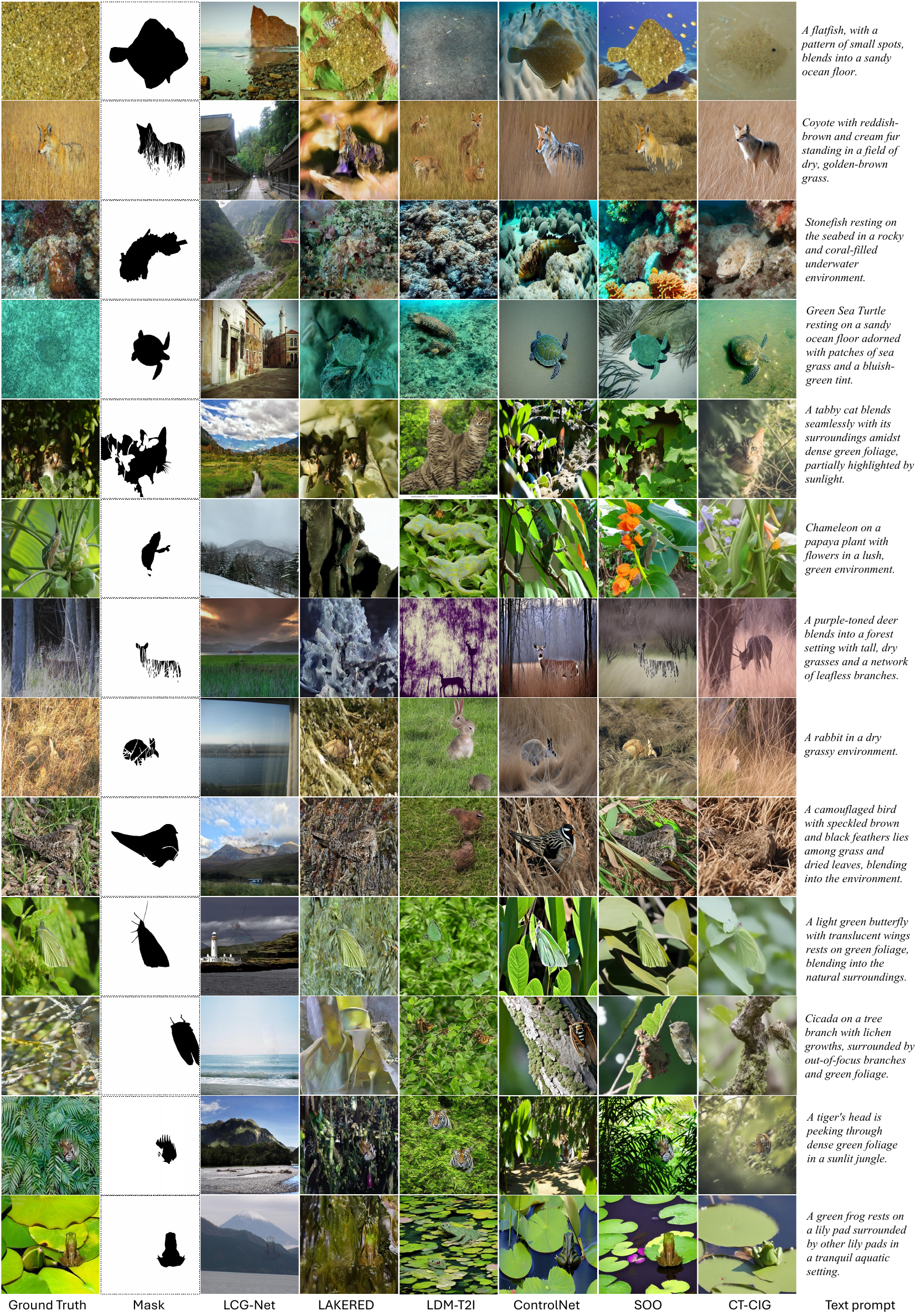}
    \caption{More results about the generated camouflage images from different methods.}
    \label{fig:more_compare}    
\end{figure*}

\begin{figure*}[t]
    \centering
    \includegraphics[width=0.9\linewidth]{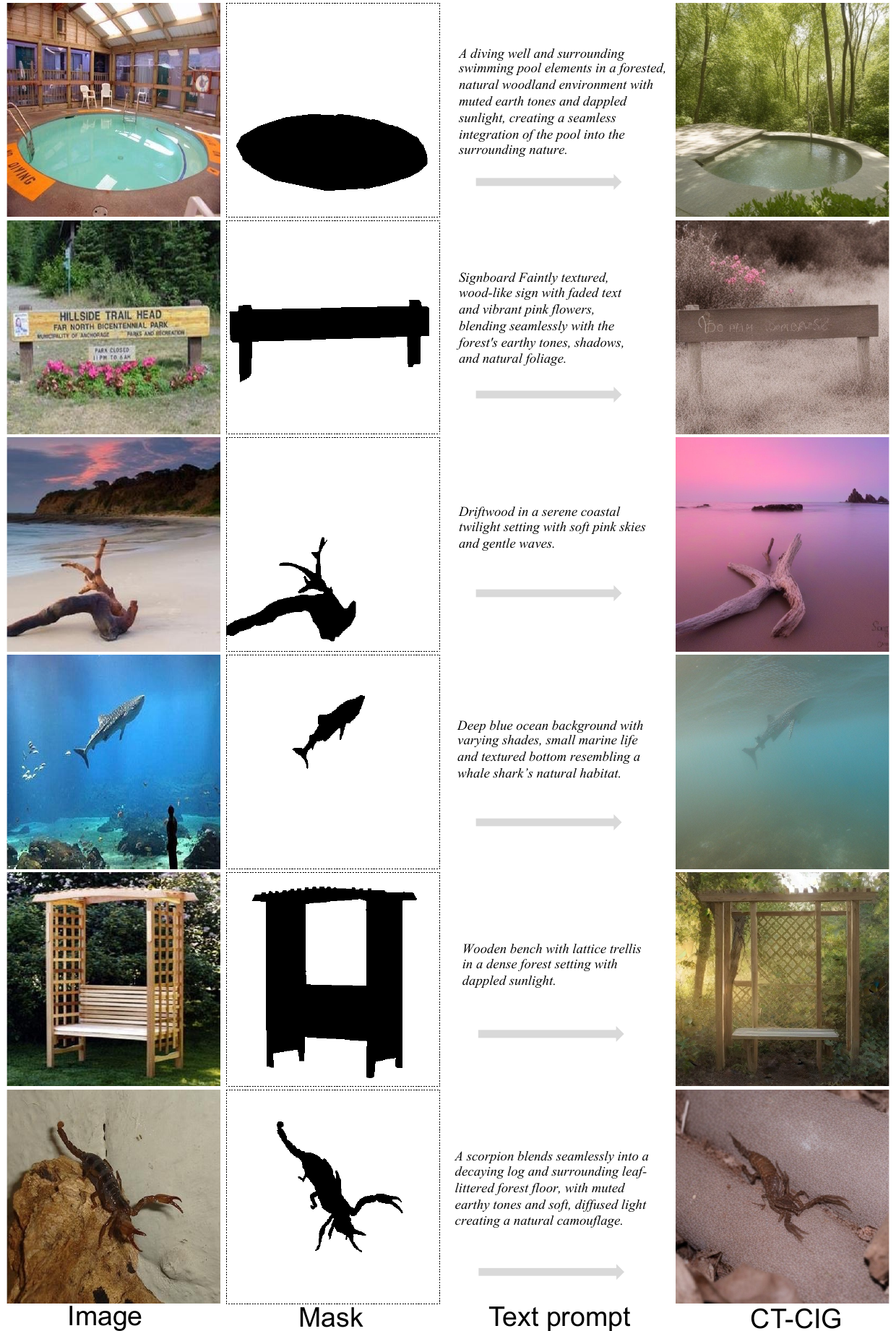}
    \caption{Camouflaged results generated by CT-CIG, taking inputs of salient objects.}
    \label{fig:sod2cam}    
\end{figure*}

\begin{figure*}[t]
    \centering
    \includegraphics[width=0.9\linewidth]{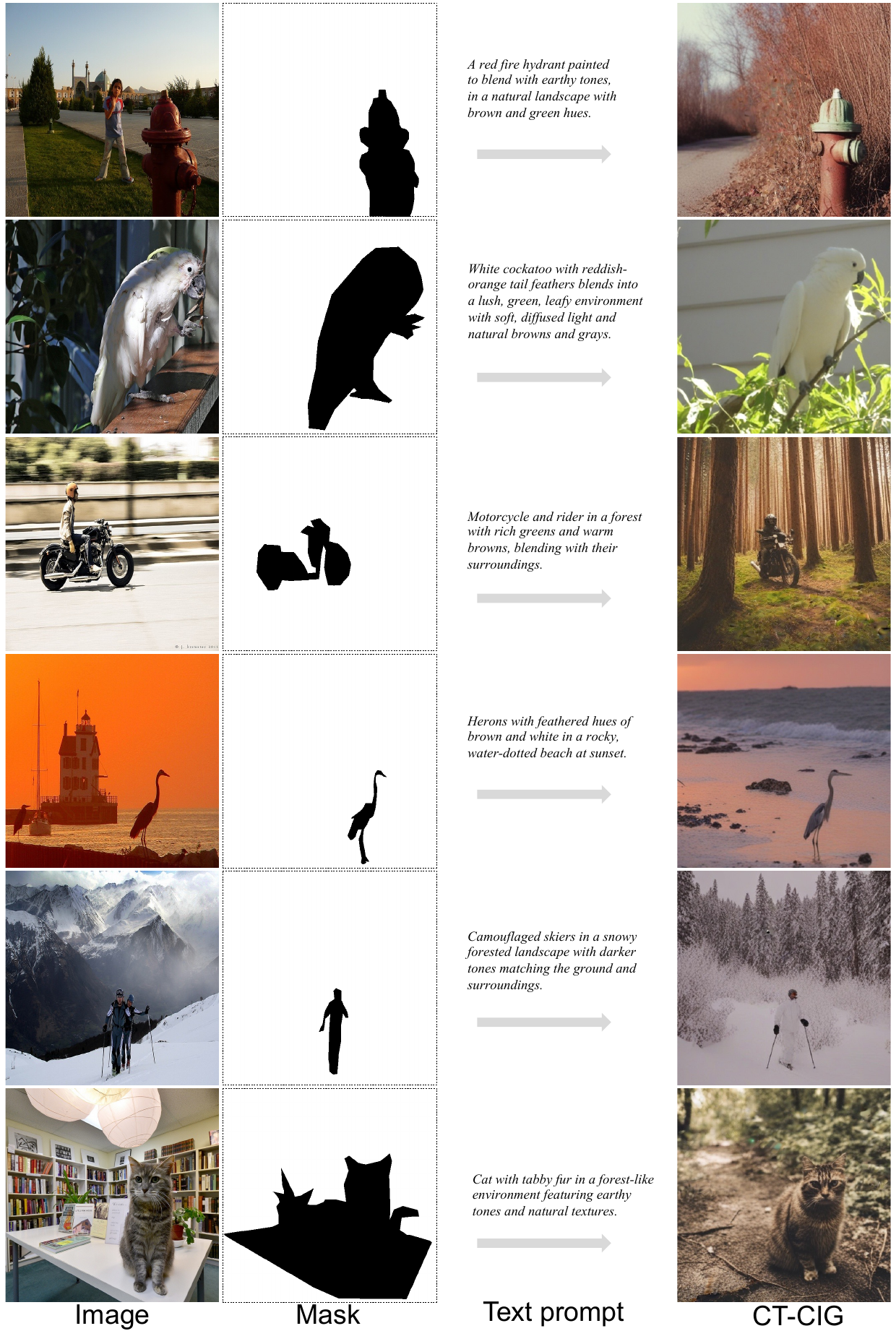}
    \caption{Camouflaged results generated by CT-CIG, taking inputs of general objects.}
    \label{fig:seg2cam}    
\end{figure*}

More camouflage images generated by CT-CIG are organized and displayed in this section due to the page restriction of the main paper.
Fig. \ref{fig:more_compare} shows the qualitative comparison of CT-CIG and other methods, taking camouflage images in COD datasets as ground truth.
Fig. \ref{fig:sod2cam} and Fig. \ref{fig:seg2cam} show the examples of using CT-CIG to transform salient object images and general images to camouflage object images, which proves the ability of CT-CIG for camouflage synthesis.

\section{D. Limitations and Future Works}
\begin{figure}[h]
    \centering
    \includegraphics[width=\linewidth]{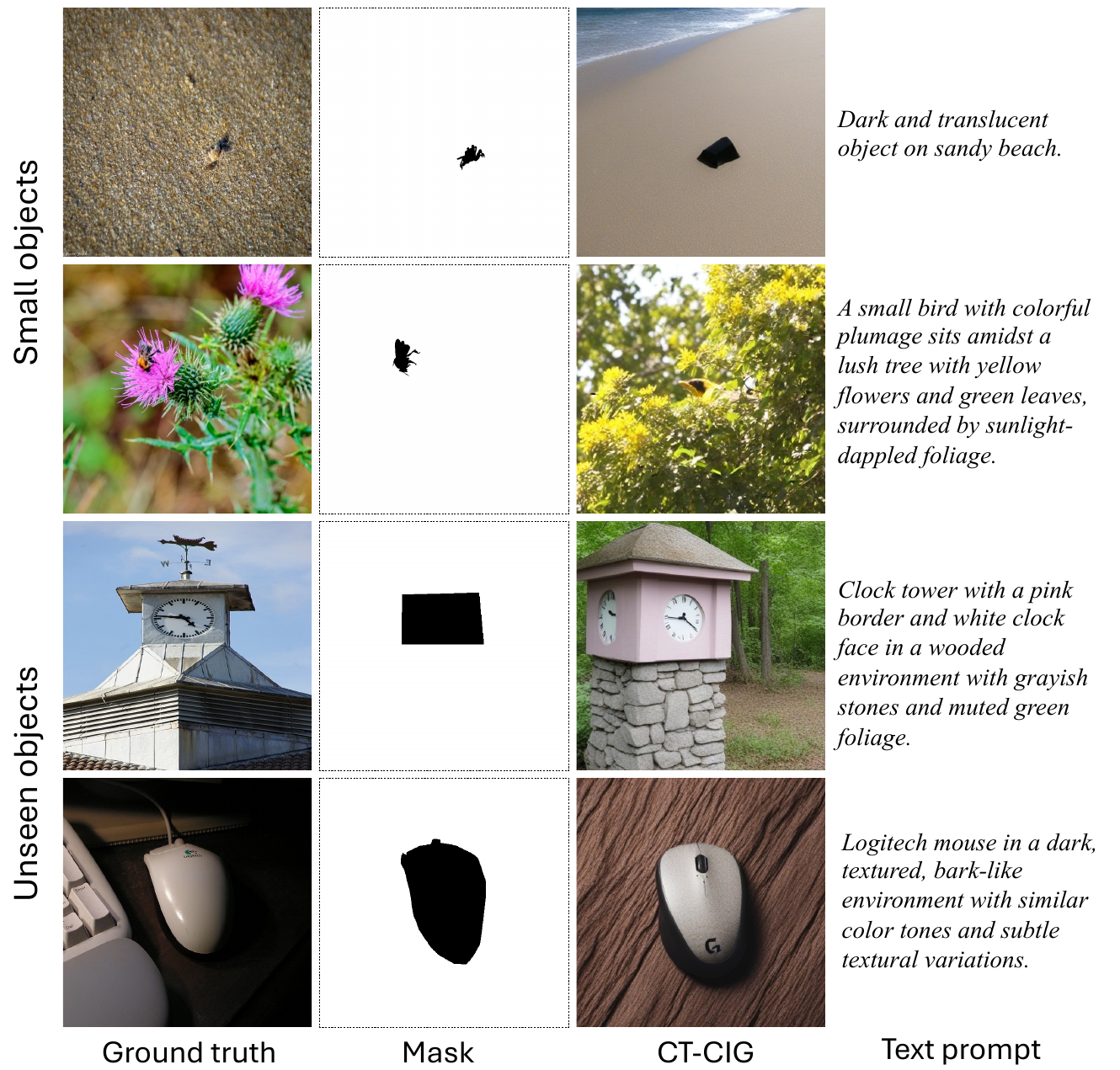}
    \caption{Failure cases of CT-CIG, which are likely to happen when meeting small objects or objects that are excluded from the training dataset.}
    \label{fig:limitation}
\end{figure}

Despite the robust capability of CT-CIG in generating realistic camouflage images, we acknowledge its limitations under specific conditions. Fig. \ref{fig:limitation} illustrates several failure cases, which can be categorized into two primary types: small objects and unseen objects (i.e., those not included in the training data).

For small objects, the failure often stems from the VLM's inability to correctly identify the target. This includes instances of complete recognition failure (e.g., Case 1) and object misclassification (e.g., Case 2). Consequently, CT-CIG fails to comprehend the logical relationship between the object and its background or the object's geometric control signals, leading to the generation of incorrect content.
Furthermore, when processing unseen objects (e.g., Cases 3 and 4), CT-CIG lacks the prior knowledge of how these objects appear in camouflaged contexts. This prevents the model from properly interpreting the intrinsic camouflage patterns described in the text prompt. As a result, the model resorts to leveraging the original SDXL parameters to generate the scene. This fallback mechanism leads to outputs that lack the intended camouflage effect and may even exhibit positional deviation from the control signal.

Our future work will focus on two primary directions: generating photorealistic camouflage images containing small objects and extending camouflage generation to object categories beyond the COD dataset.
On one hand, we plan to explore finetuning techniques to enhance the VLM's comprehension under camouflage scenarios to improve the accuracy of text prompt generation, especially for small or occluded objects. The pioneering works MM-CamObj \cite{ruan2025mm2} and MMCSBench \cite{zhangmmcsbench2} provide a promising foundation for this research. On the other hand, we will expand our data collection efforts to include a wider variety of object categories. By enriching the training set, we aim to support the generation of camouflage images for a broader spectrum of subjects, not limited to organisms found in nature. 

\end{document}